\documentclass[10pt,twocolumn,letterpaper]{article}

\usepackage{cvpr}              

\usepackage[dvipsnames]{xcolor}

\definecolor{cvprblue}{rgb}{0.21,0.49,0.74}
\usepackage[pagebackref,breaklinks,colorlinks,citecolor=cvprblue]{hyperref}

\def\paperID{13} 
\def\confName{CVPR}
\def\confYear{2024}

\title{Finding Patterns in Ambiguity: Interpretable Stress Testing in the Decision~Boundary}

\author{Inês Gomes\\ines.gomes@fe.up.pt
\and Luís F. Teixeira\\luisft@fe.up.pt
\and Jan N. van Rijn\\j.n.van.rijn@liacs.leidenuniv.nl
\and Carlos Soares\\csoares@fe.up.pt
\and André Restivo\\arestivo@fe.up.pt
\and Luís Cunha\\up201706736@fe.up.pt
\and Moisés Santos\\mrsantos@fe.up.pt
}

\begin{document}
\maketitle
\begin{abstract} 

The increasing use of deep learning across various domains highlights the importance of understanding the decision-making processes of these black-box models. Recent research focusing on the decision boundaries of deep classifiers, relies on generated synthetic instances in areas of low confidence, uncovering samples that challenge both models and humans. We propose a novel approach to enhance the interpretability of deep binary classifiers by selecting representative samples from the decision boundary --- prototypes --- and applying post-model explanation algorithms. We evaluate the effectiveness of our approach through 2D visualizations and GradientSHAP analysis. Our experiments demonstrate the potential of the proposed method, revealing distinct and compact clusters and diverse prototypes that capture essential features that lead to low-confidence decisions. By offering a more aggregated view of deep classifiers' decision boundaries, our work contributes to the responsible development and deployment of reliable machine learning systems.\footnote{\url{https://github.com/inesgomes/db-patterns}}

\end{abstract}
    
\section{Introduction} \label{sec:intro}

Nowadays, Deep Learning (DL) models are broadly used in various domains, but their lack of interpretability due to their black-box nature poses a significant challenge~\cite{BarredoArrieta2020ExplainableAI}. Recent efforts explore DL models' decision-making processes, particularly around decision boundaries, where models often struggle to make correct predictions. 
Research initiatives such as DeepDIG~\cite{Karimi2022DecisionNetworks}, GASTeN~\cite{Cunha2023GASTeN:Networks} and AmbiGuess~\cite{Weiss2023GeneratingTesting} study the decision boundary in a data-driven way by generating borderline instances, \ie synthetic low-confidence examples, using techniques like Generative Adversarial Networks (GANs) or Variational Auto-encoders (VAEs).
While many borderline instances consist of noisy data with patterns undetectable to the human eye, similar to adversarial examples~\cite{szegedy2013intriguing}, prior work has shown that a well-selected subset of such samples resembles genuinely hard-to-classify images, even for humans~\cite{Cunha2023GASTeN:Networks,Weiss2023GeneratingTesting}.
Building on this, we propose a novel approach to enhance the interpretability of deep binary classifiers by selecting representative samples from the decision boundary --- prototypes --- and applying post-model explanation algorithms. 

\begin{figure}[t]
  \centering   \includegraphics[width=0.9\linewidth]{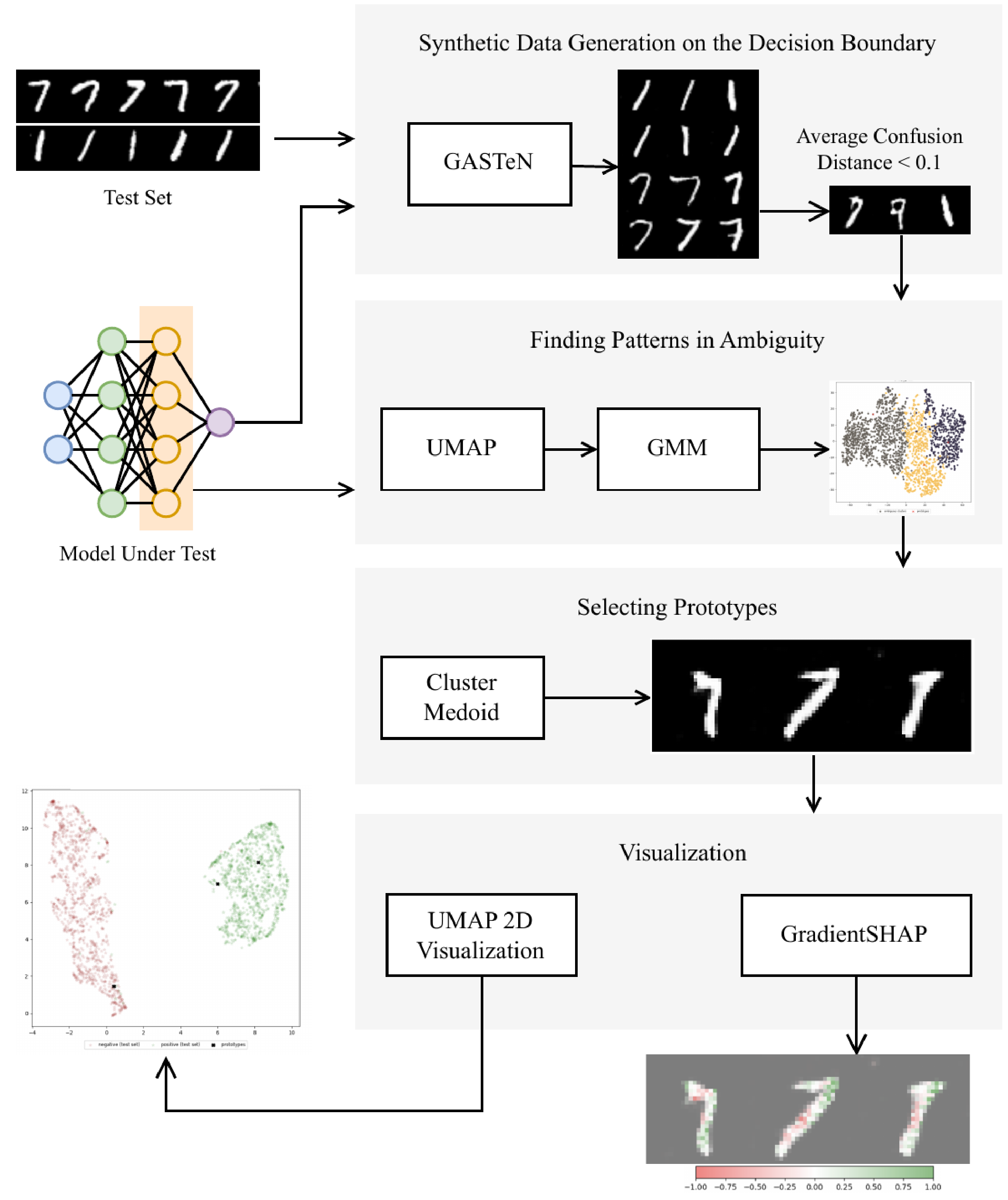}
   \caption{Schematic overview of the proposed method for improving the decision boundary interpretability of the Model Under Test by combining synthetic image generation and deep clustering.}
    \label{fig:gasten-pat}
\end{figure}

Our method, illustrated in \cref{fig:gasten-pat}, comprises four steps: 1. generate synthetic data near the decision boundary with GASTeN; 2. detect patterns in these examples using UMAP~\cite{McInnes2018UMAP:Reduction} and Gaussian Mixture Models (GMM);
3. choose a representative prototype from each cluster; and 4. visualize the prototypes and the decision boundary using 2D space visualization and GradientSHAP~\cite{Lundberg2017APredictions}. We empirically evaluated the method using three Convolutional Neural Networks (CNN) of different complexity on binary subsets of MNIST and Fashion-MNIST. Results show the potential of the method, revealing distinct, compact clusters and diverse prototypes that embody the features contributing to low-confidence decisions.

Ultimately, our method aims for a more responsible use of AI models by supporting development and auditing. During development, it can spot potential model limitations by identifying and explaining key examples the model struggles with. Labelling and using these examples for further training can improve the model through active learning or data augmentation. These examples are also valuable for developing models with a reject option~\cite{Hendrickx2021MachineSurvey} --- good models should refrain from predicting on these prototypes. Moreover, prototypes can support deployment by providing information about the data types for which a model is expected to make low-confidence predictions, serving as a semi-automatic tool to generate model cards~\cite{Mitchell2019ModelReporting}.

\section{Related Work}
\subsection{Stress Testing Machine Learning Models} 

Stress testing is an evaluation process to assess system robustness, limitations, and overall performance under challenging conditions. When applied to ML models, it involves testing models on adverse conditions, including out-of-distribution~\cite{Henriksson2019PerformanceNetworks}, adversarial~\cite{Prabhu2023LANCE:Images,Dunn2019AdaptiveInputs} or ambiguous inputs solely for the model~\cite{Cunha2023GASTeN:Networks,Liu2022DeepBoundary:Representation} or both the model and humans~\cite{Weiss2023GeneratingTesting}. Recent work has explored model decision boundaries to understand the limits of ML models. \citet{Weiss2023GeneratingTesting} generate ambiguous data points to train and test DNN supervisors; \citet{Heo2019KnowledgeBoundary} use samples near the boundary for knowledge distillation; \citet{Liu2022DeepBoundary:Representation} and \citet{Demir2023DistributionNetworks} study the decision boundary to comprehend models on safety-critical fields.

When considering only data-driven approaches, techniques such as DeepDIG~\cite{Karimi2022DecisionNetworks} and DeepBoundary~\cite{Liu2022DeepBoundary:Representation} generate adversarial examples combined with binary search to find the closest points to the decision boundary; AmbiGuess~\cite{Weiss2023GeneratingTesting} leverages autoencoders to target specific latent space distributions; GASTeN~\cite{Cunha2023GASTeN:Networks} introduces a GAN-based methodology that incorporates the output of a classifier as part of the generator's loss function; and \citet{Demir2023DistributionNetworks} combine state-of-the-art methods, including image transformations, GANs and adversarial attacks, followed by ML models that select those with highest uncertainty.

Few studies explore how to use samples that fall within a model’s low-confidence region for responsible AI. \citet{Demir2023DistributionNetworks} suggest using post-model explanations on error-prone class samples, while \citet{Cunha2023GASTeN:Networks} incorporates such samples into model cards.
\subsection{Slice Discovery Methods}

Slice discovery consists of methods that identify semantically meaningful subgroups within unstructured data, particularly where models perform poorly~\cite{Eyuboglu2022Domino:Embeddings}. Our approach, while similar, diverges by focusing on low-confidence instances. These methods leverage deep clustering, a technique that uses neural networks to capture relevant input features, followed by traditional clustering algorithms~\cite{Min2018AArchitecture}.

The slice discovery methods state-of-the-art show numerous techniques for representing image data. GEORGE~\cite{Sohoni2020NoProblems} and PlaneSpot~\cite{Plumb2023TowardsModels} extract the embeddings from the penultimate layer of their model, whereas DOMINO~\cite{Eyuboglu2022Domino:Embeddings} uses pre-trained embeddings, more specifically CLIP~\cite{Radford2021LearningSupervision} and ConVIRT~\cite{Zhang2022ContrastiveText}. These methods then fine-tune GMMs for clustering.  

These methods employ dimensionality reduction when facing challenges with high-dimensional data, such as inefficient similarity measures~\cite{Min2018AArchitecture} and the curse of dimensionality. DOMINO uses PCA with 128 components, PlaneSpot opts for \textit{scvis}~\cite{Ding2018InterpretableModels} with 2 dimensions, and GEORGE selects UMAP with 1 or 2 components based on the dataset. Given that UMAP helps preserve the essential structure of the data, combining UMAP with GMM, as seen in studies like N2D~\cite{McConville2020N2D:Embedding}, demonstrates that manifold learning techniques can significantly improve clustering quality by considering the local data structure.
\section{Borderline Prototype Generation }\label{sec:method} 

\Cref{fig:gasten-pat} summarizes the proposed method. The model that is being subjected to the stress-test must be a deep binary image classification, and it is referred to as the Model Under Testing (MUT). First, we populate the decision boundary by generating synthetic images close to the MUT decision boundary, \ie our borderline instances. To that end, we employ GASTeN, a GAN-based technique trained with the MUT predictions, to approximate its decision boundary~\cite{Cunha2023GASTeN:Networks}. We chose GASTeN as it generates realistic challenging borderline examples for a specific classifier. Then, we filter the synthetic images using the Average Confusion Distance (ACD) \citet{Cunha2023GASTeN:Networks}, that measures the closeness of a sample to the decision boundary. We filter the images by $\mathit{ACD} < 0.1$ to ensure only low-confidence predictions. We assess the quality of the borderline images by calculating the Fréchet Inception Distance (FID) scores~\cite{Heusel2017GANsEquilibrium}. 

In our second step, we apply deep clustering to find patterns in the borderline instances generated in the previous step. To that end, we extract the high-level feature embeddings from the MUT's penultimate layer. 
Then, we apply UMAP for dimension reduction, followed by GMM clustering to group visually similar images. We selected this combination given the favourable findings in the literature review~\cite{Sohoni2020NoProblems,Plumb2023TowardsModels,Eyuboglu2022Domino:Embeddings}. Particular hyperparameters are tuned, considering the specific characteristics of the low-confidence region, as we explain in \cref{sec:model-archi}.
Finally, the quality of the resulting clusters is assessed through the silhouette score~\cite{Rousseeuw1987Silhouettes:Analysis} and the Davies-Bouldin index~\cite{Davies1979AMeasure}. We use these measures of cluster definition and separation to ensure the formation of distinct and coherent groups.

In the third step, we select the medoid from each cluster to represent the cluster. 
The medoid is calculated by minimizing the sum of distances to all other objects in that cluster. As a centrally located sample, it ensures a robust representation of each identified pattern.

In the fourth step, we evaluate the representativeness of the selected prototypes through visual inspection. Our goal is to generate prototypes that demonstrate greater feature diversity, more dispersed distribution across the 2D space, and enhanced interpretability through GradientSHAP maps. With this in mind, we train UMAP on the test set to capture the structure of the original data. Then, we analyze the positioning of the prototypes within the 2D space created. For an in-depth analysis of why these images are close to the decision boundary, we use GradientSHAP --- a technique that explains the contribution of each pixel to the model's output by integrating gradients with SHAP values~\cite{Lundberg2017APredictions}.

\section{Experimental Setup}
%
%
\subsection{Dataset}

We use MNIST~\cite{LeCun1998Gradient} and Fashion-MNIST~\cite{Xiao2017Fashion-mnist:Algorithms} datasets to evaluate our method. We chose these datasets for their interpretability without needing expert knowledge, simplicity in size, and lack of color. We created binary subsets from these datasets for binary classification, focusing on similar concepts: \emph{7 vs 1}, \emph{8 vs 0} and \emph{5 vs 3} for MNIST and \emph{dress vs top} and \emph{sneaker vs sandal} for Fashion-MNIST.
\subsection{Model Architecture}\label{sec:model-archi}

To obtain more general conclusions, we evaluated MUTs architectures of varying complexities. Following the GASTeN study, we utilized a CNN architecture with two convolutional blocks, where the complexity is adjusted by varying the number of filters~\cite{Cunha2023GASTeN:Networks}. 
For the MNIST dataset, we tested CNN models with 1, 2, and 4 filters, while for Fashion-MNIST, we used 4, 8, and 16 filters.

To train GASTeN, we adapted its setup based on previous findings by \citet{Cunha2023GASTeN:Networks}, tailoring its hyperparameters towards our stress-testing objectives. Training GASTeN required choosing two specific hyperparameters beyond the standard DCGAN parameters: the confusion distance weighting ($\alpha$) and the pre-training epochs. The $\alpha$ value critically influences GASTeN's loss function, while the pre-training duration affects image realism. Based on the original authors' suggestions, we opted for 5 pre-training epochs for MNIST and 10 for Fashion-MNIST, each with an $\alpha$ weight of 25. We determined GASTeN's optimal training duration by optimizing the FID-ACD minimization~\cite{Cunha2023GASTeN:Networks}, leading to selecting 10 epochs for MNIST and 15 for Fashion-MNIST. We generated 15,000 synthetic images for each task.

For deep clustering, we optimized the silhouette score using the Bayesian hyperparameter optimization method with 25 iterations. With UMAP, we investigated the optimal number of neighbours to balance local versus global data structures. Given our focus on classifying similar concepts and analyzing regions of low confidence, where features are less distinct, our analysis prioritized local structures. Therefore we explore a range of 5 to 25 neighbours. We also varied the minimum distance between 0.01 and 0.25 to control embedding compactness and set the components between 10 and 60 to ensure detailed clustering without losing critical information. For GMM, we varied the cluster count from 3 to 15 to maintain a practical number of prototypes for analysis. We also selected the covariance as \textit{full}, as it allows for each cluster to have its covariance matrix. 

\section{Results and Discussion}
\subsection{Synthetic Data Generation}

We tested three MUTs on the five binary subsets. 
The MNIST \emph{5 vs 3} subset using a CNN with one filter attained the lowest accuracy of 92.53\%, while the \emph{8 vs 0} subset with a four-filter CNN reached the highest accuracy of 98.92\%.

After training GASTeN for each classifier-dataset subset combination, we generated 15,000 synthetic images and subsequently filtered those with $\mathit{ACD < 0.1}$. This process resulted in an average FID score increase of 250 points for images near the decision boundary and an average 86\% reduction in image count post-filtering. The significant FID score rise and the substantial image count decrease after filtering suggest GASTeN's limited efficiency in generating decision boundary-near samples. 

During this process, we observed some correlation between the model complexity and GASTeN FID scores ($\mathit{\rho_{nf, FID}= 0.52}$). We expected this outcome, as lower classifier capacities lead to more challenging classifications, resulting in less confident images. However, a contrary example is our least accurate classifier, which produced realistic (low FID scores) synthetic images that were unrealistic (high FID scores) near the decision boundary. 
\subsection{Finding Patterns in Ambiguity}

With the resulting synthetic images from the previous step, we optimized UMAP and GMM hyperparameters to achieve the highest silhouette score. UMAP frequently select hyperparameters that highlight the local structure and preserve image ambiguity. GMM clustering resulted in either a small (\eg~3) or large (\eg~15) number of clusters without a discernible pattern.

Evaluation metrics indicated modest clustering quality, including the silhouette score (0.26 --- 0.52) and the Davies-Bouldin Index (0.7 --- 1.35) on the 15 classifier-dataset subset pairs. These metrics suggest that while clusters are reasonably distinct and compact. There is room for improvement, possibly due to some overlap or sparseness in clusters. The best performance occurred on the MNIST \emph{7 vs 1} subset with a four-filter CNN and the poorest on the \emph{5 vs 3} subset with a four-filter CNN. 

We noticed a negative correlation between the quantity of low-confidence images and the silhouette score ($\mathit{\rho_{\# images, SIL}= -0.62}$), indicating that fewer images generally lead to more effective clustering. We suspect this could be due to noise in the generated images, suggesting that exploring alternative dimensionality reduction and clustering techniques could enhance our clustering results.
\subsection{Selecting and Visualizing Prototypes}

After clustering, we select the medoid of each cluster as our prototype and assess its representativeness through a 2D visualization and with GradientSHAP. 
To illustrate the usefulness of the proposed method, we chose the best-performing subset based on the silhouette score for this section analysis: MNIST subset \emph{7 vs 1} with four-filter CNN.

\begin{figure}
  \centering
  \begin{subfigure}{0.495\linewidth}
    \includegraphics[width=\linewidth]{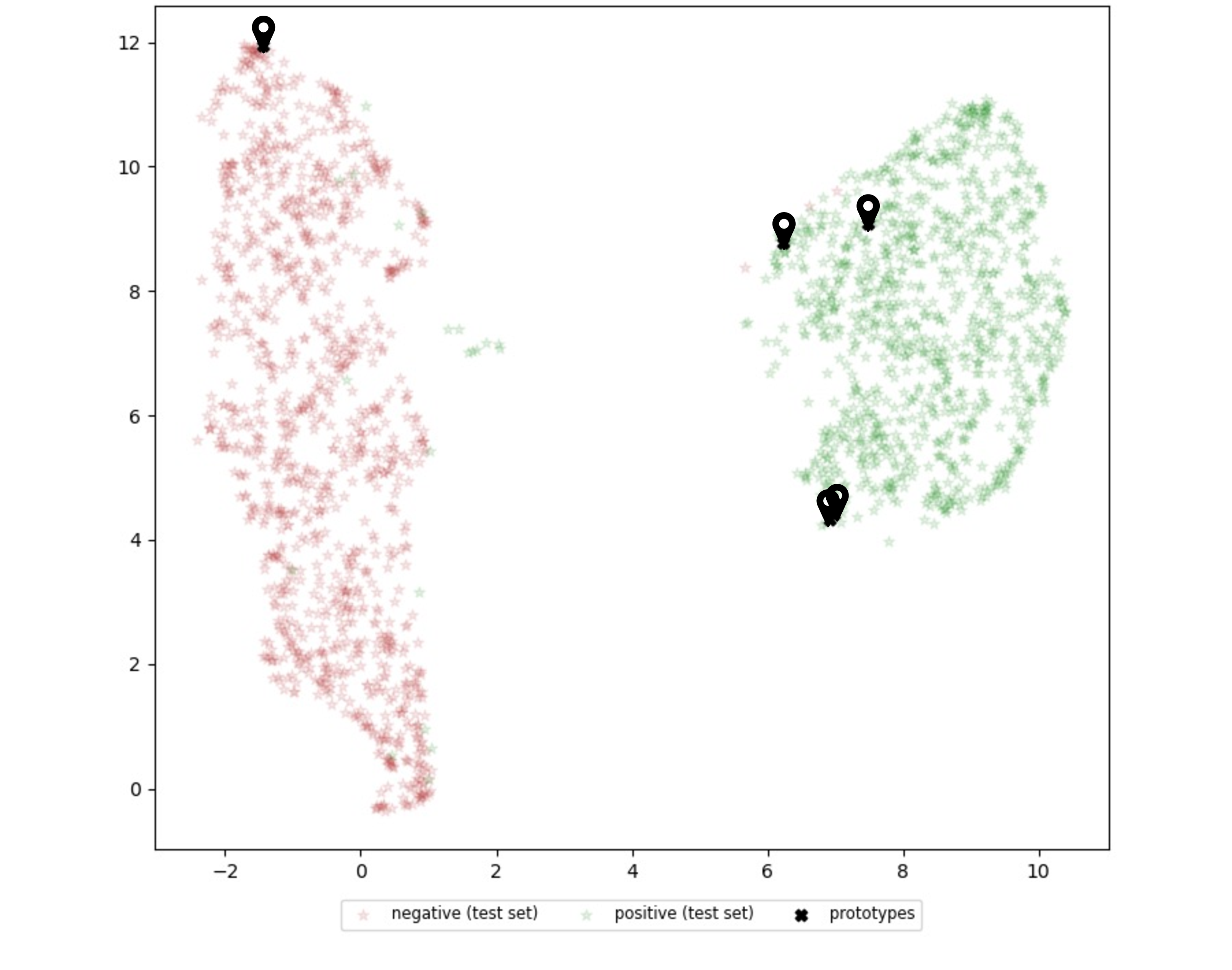}
    \caption{Baseline}
    \label{subfig:2dbasemnist}
  \end{subfigure}
  \hfill
  \begin{subfigure}{0.495\linewidth}
    \includegraphics[width=\textwidth]{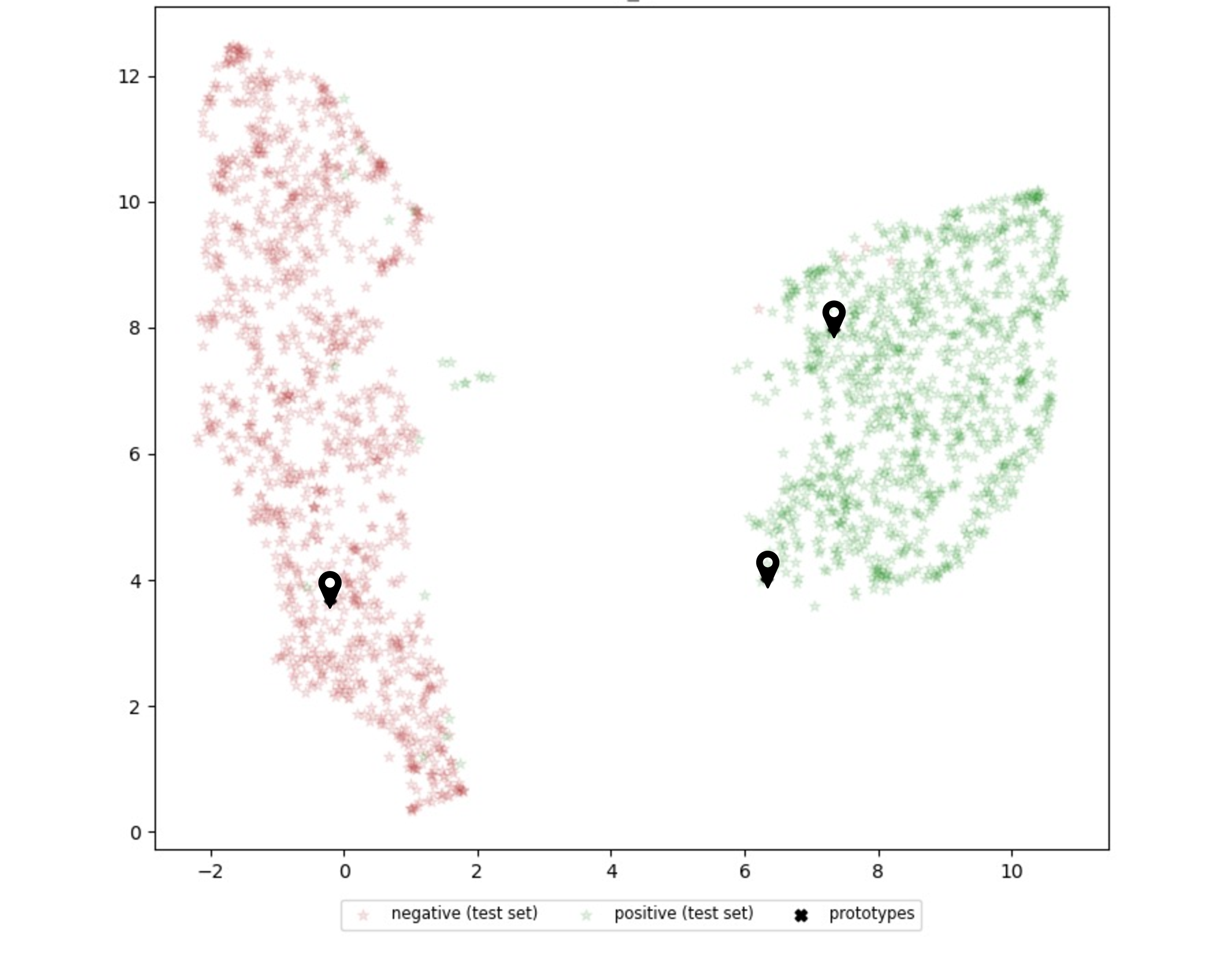}
    \caption{Prototypes}
    \label{subfig:2dprotomnist}
  \end{subfigure}
    \caption{UMAP 2D space for MNIST \emph{7 vs 1} and four-filter CNN. The test set is marked by stars with \emph{1} in red, and \emph{7} in green. Black pins indicate the prototypes or baseline positions.}
  \label{fig:2dviz}
\end{figure}

\Cref{fig:2dviz} shows the distribution of prototypes versus the baseline in the UMAP 2D space. In this example, we conclude that prototypes are more dispersed than the baseline which even includes overlapping images. 

\begin{figure}
  \centering
  \begin{subfigure}{0.55\linewidth}
    \includegraphics[width=\linewidth]{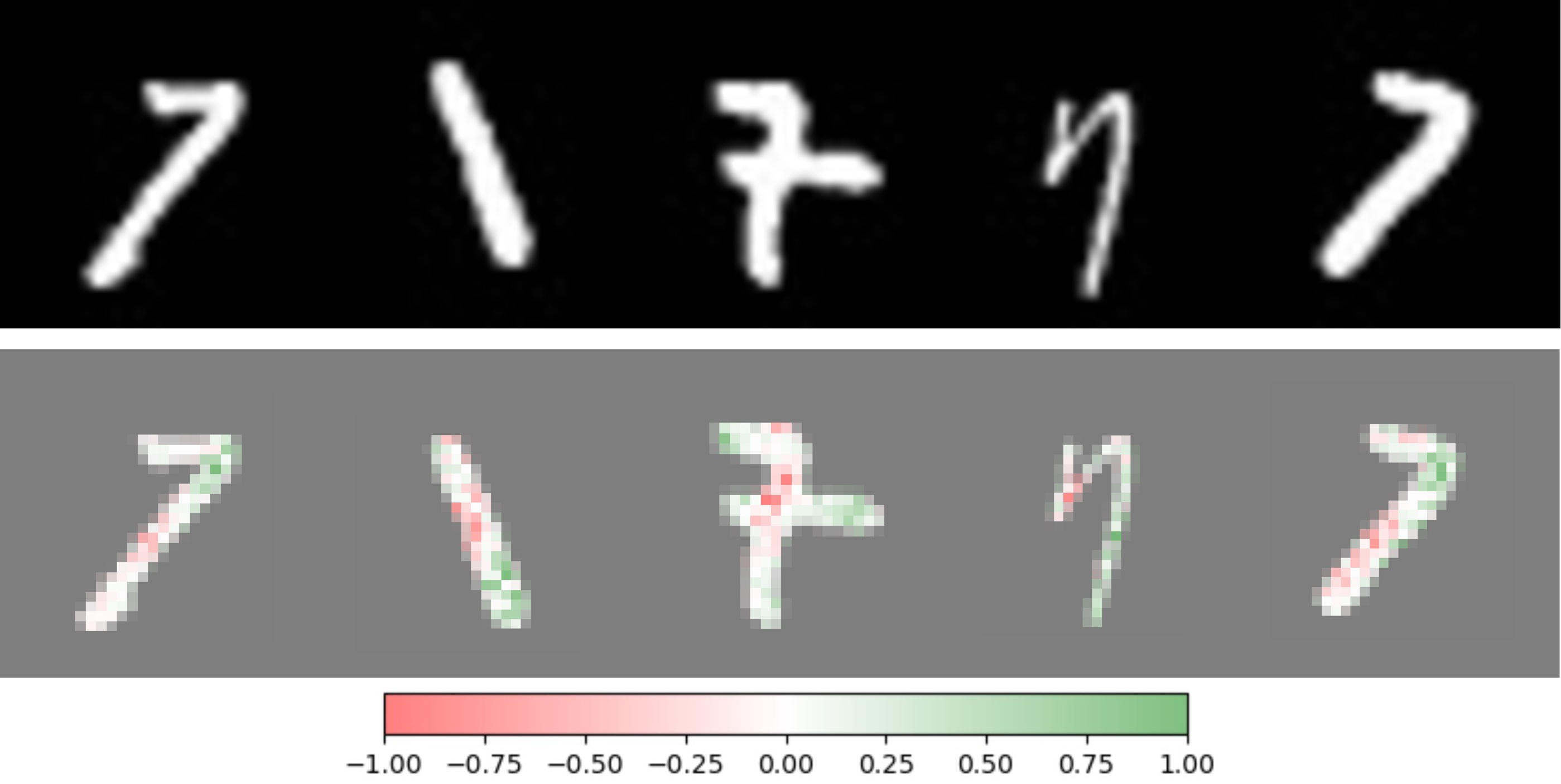}
    \caption{Baseline}
    \label{subfig:basemnnist}
  \end{subfigure}
  \hfill
  \begin{subfigure}{0.33\linewidth}
    \includegraphics[width=\textwidth]{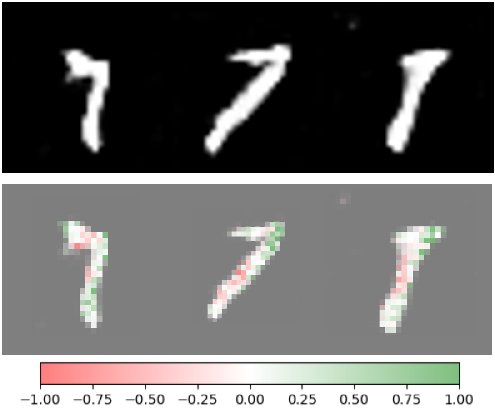}
    \caption{Prototypes}
    \label{subfig:protomnist}
  \end{subfigure}
    \caption{Selected images and the corresponding GradientSHAP maps for MNIST \emph{7 vs 1} and four-filter CNN. Features contributing to the classification of \emph{1} are red, and \emph{7} are green.}
  \label{fig:gradients}
\end{figure}

Observing the baseline in \cref{subfig:basemnnist}, two images appear remarkably similar, likely those clustered together in the 2D space, with a fourth image resembling noise. In contrast, in \cref{subfig:protomnist}, the prototypes display distinct features, particularly regarding rotation. We also note that all images lack part of the seven's upper section. We hypothesize that this feature is an intrinsic attribute of the low-confidence region in this MUT, and that the prototypes effectively capture it.

GradientSHAP maps indicate that features leading to classification as the positive class (7) include the top elbow and the middle line of the seven. On the other hand, attributes favoring the negative class (1) typically relate to pixels in the center of the image. Remarkably, these maps also express uncertainty regarding the missing portion of the seven's upper section, validating our hypothesis that the model has difficulty learning this feature.

\section{Conclusions}

In this work, we investigate borderline instances that contain visual properties that make predictions complex, even for humans. We study the impact of combining synthetic image generation and deep clustering on the interpretability of deep binary classifiers' decision boundary. 

Further research includes improving the generation of ambiguous images, exploring other clustering and embedding techniques for improved performance, and developing quantitative metrics to validate our prototypes statistically. Additionally, as we tested our approach on simplified datasets, where it is possible to assess ambiguity visually, the performance in complex scenarios needs to be assessed.

Nevertheless, we obtained promising results that show that it is possible to uncover patterns in borderline images. By visual inspection, we can study the representativeness of our prototypes and the features associated with the low-confidence region. Such insights are invaluable for auditing machine learning models, identifying and mitigating potential weaknesses during development, or documenting the limitations of classifiers in model cards upon deployment.
\paragraph{Acknowledgments} AISym4Med (101095387) through the Horizon Europe Cluster 1: Health, ConnectedHealth (n.º 46858); Competitiveness and Internationalisation Operational Programme (POCI) and Lisbon Regional Operational Programme (LISBOA 2020), under the PORTUGAL 2020 Partnership Agreement, through the European Regional Development Fund (ERDF); NextGenAI - Center for Responsible AI (2022-C05i0102-02), supported by IAPMEI; FCT plurianual funding for 2020-2023 of LIACC (UIDB/00027/2020 UIDP/00027/2020).

{
    \small
    \bibliographystyle{ieeenat_fullname}
    \bibliography{references}
}



\end{document}